\documentclass[11pt,a4paper]{article}

\usepackage[hyperref]{naaclhlt2019}
\usepackage{times}
\usepackage{latexsym}
\usepackage{amssymb}
\usepackage{mathtools}
\usepackage{url}
\usepackage{graphicx}
\usepackage{caption}
\usepackage{subcaption}
\usepackage{verbatim}
\usepackage{algorithm}
\usepackage{tcolorbox}
\usepackage{color}
\usepackage{soul}
\usepackage[inline]{enumitem}
\usepackage{stmaryrd}
\usepackage{arydshln}
\usepackage{upgreek}

\usepackage{enumitem}
\usepackage{natbib}
\usepackage{mathtools}

\usepackage{algorithmicx}
\usepackage{algpseudocode}

\aclfinalcopy % Uncomment this line for the final submission

\usepackage{xcolor}
\newcommand{\ignore}[1]{}

\title{Improving Query Safety at Pinterest}

\author{Abhijit Mahabal, Yinrui Li, Rajat Raina, Daniel Sun, Revati Mahajan, Jure Leskovec\\
Pinterest \\ {\tt amahabal@, yinruili@, rraina@, dsun@, rmahajan@, jure@} \\
{\tt pinterest.com}}

\date{June 2020}

\begin{document}
\newcommand{\todo}[1]{{[\scshape Todo: #1]}}
\newcommand{\query}[1]{\emph{[#1]}}
\newcommand{\jure}[1]{{\color{red}[J: #1]}}
\newcommand{\hide}[1]{}
\newcommand{\xhdr}[1]{\vspace{2mm}{\noindent\bfseries #1}.}
\newcommand{\xhdrp}[1]{\vspace{2mm}{\noindent\bfseries #1}}

\maketitle

\begin{abstract}
	Query recommendations in search engines is a double edged sword, with undeniable benefits but potential of harm. Identifying unsafe queries is necessary to protect users from inappropriate query suggestions.  However, identifying these is non-trivial because of the linguistic diversity resulting from large vocabularies, social-group-specific slang and typos, and because the inappropriateness of a term depends on the context.
	Here we formulate the problem as query-set expansion, where we are given a small and potentially biased seed set and the aim is to identify a diverse set of semantically related queries.
	We present PinSets, a system for query-set expansion, which applies a simple yet powerful mechanism to search user sessions, expanding a tiny seed set into thousands of related queries at nearly perfect precision, deep into the tail, along with explanations that are easy to interpret. PinSets owes its high quality expansion to using a hybrid of textual and behavioral techniques (i.e., treating queries both as compositional and as black boxes).
	Experiments show that, for the domain of drugs-related queries, PinSets expands 20 seed queries into 15,670 positive training examples at over 99\% precision. The generated expansions have diverse vocabulary and correctly handles words with ambiguous safety. 
	PinSets decreased unsafe query suggestions at Pinterest by 90\%.
\end{abstract}

\section{Introduction}

Several practical tasks require the identification of queries in a particular domain. To pick three examples of topics at different granularities, we may wish to obtain queries about travel to Paris, about travel in general, or about drugs. Large collections of such queries in a domain can be used in multiple fashions to build classifiers, for ad targeting, for blacklisting, or for enabling users to explore nearby parts of a semantic space.

One particularly important domain and an accompanying practical task is the identification of unsafe queries and the sanitizing of query suggestions. Query recommendations in search engines can have big positive as well as negative effects. It has undeniable benefits, since the suggestions not only save typing, more crucially, they enable exploration and discovery, introducing users to unanticipated but relevant regions of the search space. However, inappropriate suggestions negate all benefits, resulting in not just a bad user experience but actual harm. The danger from bad completions is real: \cite{Baker2013} shows how Google's query completions were perceived as racist, sexist, or homophobic.

\subsection{Challenges in Identifying Topical Queries}

The technical difficulties involved in obtaining queries has four big contributors: gigantic vocabulary, ambiguity, platform specificity, and lopsided data. 

\xhdr{The Challenge from Gigantic Vocabularies} In any given unsafe category (say, porn or illegal drugs) there is a vast diversity of how people refer to these topics, arising from the real underlying diversity of content but further enlarged by slang, social variability in argot, and typos. For example, there are hundreds of illegal drugs, and marijuana alone has many dozen names including \emph{pot, dope, ganja, grass, mary jane, reefer} and \emph{weed}. Likewise, queries about Paris travel involve names of attractions or of Paris neighborhoods.

\begin{figure*}
	\includegraphics[width=0.47\textwidth]{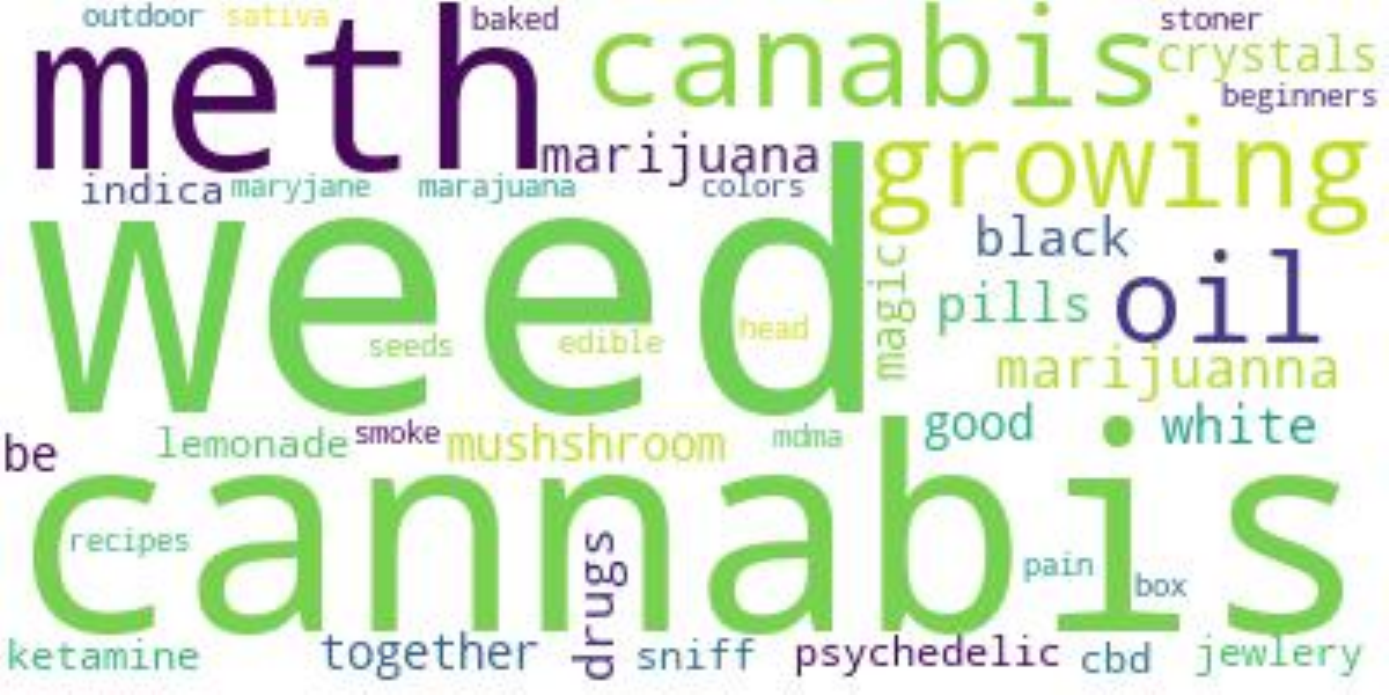}\hfill
	\includegraphics[width=0.47\textwidth]{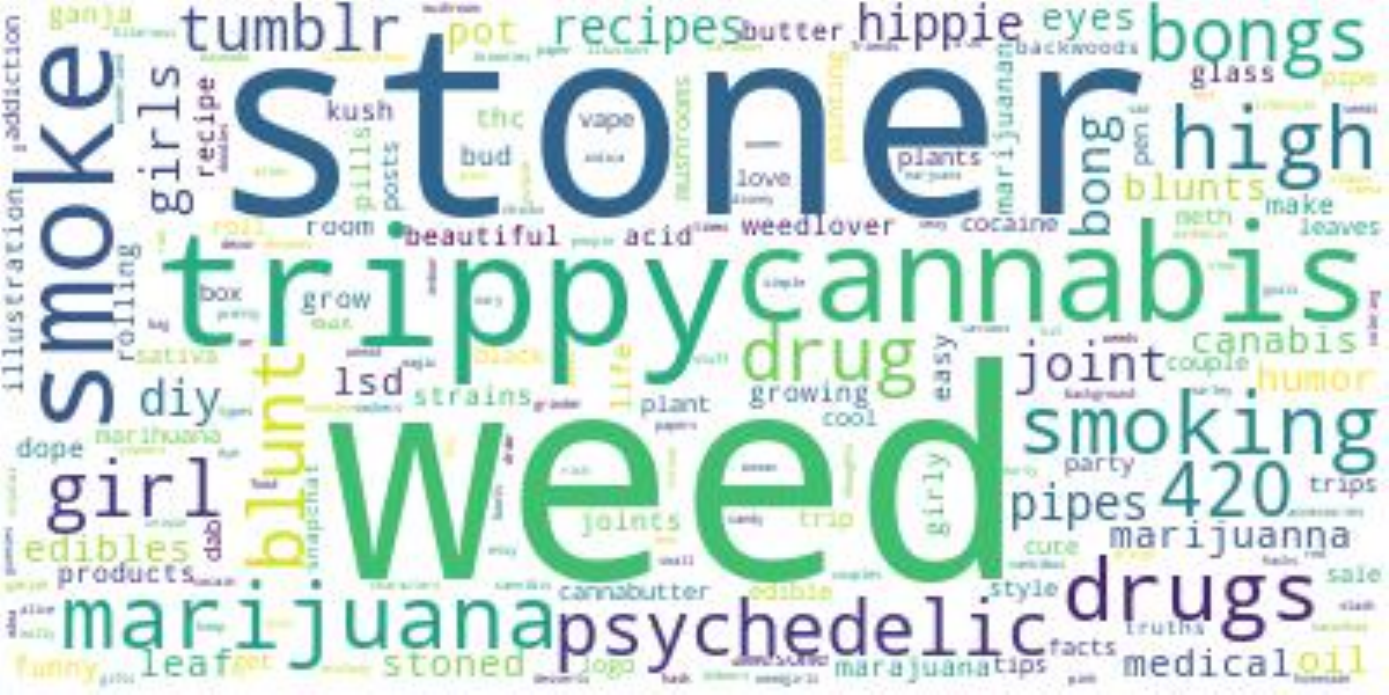}
	\caption{Word clouds of queries relating to drugs. The left side shows 20 seed queries. PinSets expanded these to 15,670 unique real queries, at 99\%+ precision, encompassing a diverse vocabulary necessary for training a robust classifier.}
	\label{fig:teaser}
\end{figure*}

\xhdr{The Challenge from Ambiguity} A single word can be either safe or unsafe, depending on the context. \emph{Grass} and \emph{weed} have perfectly safe meanings, and even words seemingly emblematic of the ``unsafe'' class, such as \emph{nude}, \emph{sexy} or \emph{adult}, have perfectly safe senses: \emph{nude lipstick, adult in the room} and \emph{sexy car}, among many.
Adding to this challenge, everyday words are routinely co-opted into unsafe senses, for example, \emph{kitty}, \emph{hole}, and \emph{beaver}.

\xhdr{The Challenge from Platform Specificity} Online platforms naturally differ in the distribution of concepts they cater to. Queries encountered by LinkedIn, Google, Pinterest, Zillow, Google Maps and Pornhub differ, as do the meanings of words and distribution of senses. This limits the utility of importing models that have been trained on a very different distribution and semantics of queries.

\xhdr{The Challenge from Lopsidedness of Real Data} Any corpus is necessarily imbalanced in various ways: languages are not equally prevalent,
individual topics are unequally represented (with a topic such as home improvements far more prevalent than the topic of axe throwing), and labeled data also unintentionally stress some topics over others. Many standard statistical techniques, such as Point-wise Mutual Information \cite{Fano1961}, place a heavy significance on even a single co-occurrence of two rare items \cite{Role2011}, and the lopsidedness adds to classification difficulties.

\xhdr{Attributes of Effective Training Data} Expanded query lists are often used for training classifiers. For a machine-learned model to be effective in the presence of such challenges as diverse vocabulary and ambiguous terms, training data must be abundant; for individual terms such as \emph{nude}, it must properly balance examples in safe vs. unsafe classes; the queries must be diverse, and make use of a wide swathe of unsafe terms, typos, and diverse phrasing; and finally, should be platform specific, tuned to the particular distribution of queries actually encountered.

\subsection{Typical Problem Formulation is Textual}

Unsafe query identification is typically formulated as a textual problem: the goal is to map a query to a safety score based solely on the text of the query and without making use of prior encounters with the query. A textual solution is both powerful and easy to deploy. The power comes from compositionality in language: the meaning of a query \query{A B C D} typically combines the meaning of its constituents. Modern machine learning techniques excel at learning compositional representations. 

However, language is not fully compositional. The safety of a query is not simply the combination of the safety of its parts. The queries \query{nude} and \query{sexy} are unsafe, but \query{nude lipstick} and \query{sexy car} aren't. In the other direction, the following queries are made up of simple and safe words but explicitly concern marijuana: \query{wake and bake} and \query{moon rock bud}.

Purely textual methods perform sub-optimally on queries that are non-compositional with the query's safety polarity reversed from that of its constituents.

\subsection{Present Work}

Here we present PinSets, a mechanism for expanding a tiny number of seed queries into two sets: a positive set with a very large number of queries of the same semantic type as the seeds and a negative set with a large number of queries not of that type. For instance, 20 seed queries concerning drugs produced over 15,000 queries about drugs at over 99.3\% precision, and produced about 1.7 million non-drug queries at almost perfect precision. The expansions cover a wide vocabulary, including typos and slang. Moreover, queries containing ambiguous words (such as \emph{pot} and \emph{weed}) are distributed appropriately among the positive and negative sets. Figure~\ref{fig:teaser} showcases the diverse terminology made accessible by PinSets.

\xhdr{Combining Textual and Behavioral} Our main insight is that we can combine textual methods (which ``look inside a query'' and understand it compositionally) with statistical behavioral methods (which consider a query as a black box, as a single, unanalyzed unit).

Black box methods provide resilience to ambiguous words. A query (say, \query{instant pot meal}) may contain an ambiguous term (here, \emph{pot}), but the term is not ambiguous in the context of the full query. When users issue the query, it is often in sessions containing recipes and cooking. We learn to associate the full query with those ngrams. Other queries connected to those ngrams will also be about cooking, notwithstanding the ambiguity in the word \emph{pot}.

The ngrams are obtained by looking within queries, and this provides us with the generalization benefits of compositional models. A query that uses completely different words, e.g., \query{pressure cooker dinner}, may perhaps never co-occur in a session with our original query, and yet share many of the ngrams. This sharing of ngrams among queries with non-overlapping terms and which don't ever co-occur in sessions allows us to expand to queries with diverse vocabulary. Even the typo-infested query \query{intsant pot mael}, if seen often, is likely to have been seen with queries about cooking and recipes.

\xhdr{Query cohesion within search sessions} A second insight may be labeled \emph{topic stickiness}. A session consists of queries issued by a single user in a short time span. Here, we restrict ourselves to small sessions with between 5 and 20 queries. For a query $q$ in a session $s$, if at least three other queries are about drugs, that is weak evidence that $q$ is about drugs. By looking at all sessions containing $q$, we can calculate the fraction of sessions that are drugs-related, providing us with a drug-relatedness estimate for $q$. For many queries, this fraction will be very high, and for many others, this will be very low, giving us confident slices of positive and negative queries. Those with a middling score (e.g., \query{high on life} and \query{club life nightclub}) are potentially ambiguous, and we do not include them in either set. 

\xhdr{Family Resemblance in Query Expansion} Ludwig Wittgenstein famously described how members of a family resemble each other although no feature is shared among all members \cite{Wittgenstein1953}. Some members may share a style of chin, others share eye color, and so forth. Queries that are about a single topic also have family resemblance: many queries about drugs co-occur in sessions with the word \emph{smoke}, a different but overlapping set of queries share \emph{ganja}, and so forth. PinSets identifies ngrams strongly associated with the seed queries, and each query in the expansion can be traced to several of these ngrams, an invaluable debugging aid. Different queries in the expansion are central or peripheral depending on how many of the selected ngrams they are linked to. In this sense, analogous to human conceptual categories, the expansion has a graded structure \cite{Lakoff2008}.

\xhdr{Resilience to Data Imbalance} The corpus consisting of all sessions naturally has an uneven distribution of topics. Also, the seed set being expanded can unintentionally stress some subset. PinSets' algorithm utilizes a few tricks to limit the effect of such imbalance, and is described at the appropriate stage.

\xhdr{Balancing Ambiguous Terms} Consider three\linebreak queries containing the term \emph{high}: \query{high eyes stoner}, \query{riverdale high}, and \query{sky high}. The first is clearly about drugs, the next clearly not, and the last is ambiguous, with a drugs-related sense and with a non-drugs sense. PinSets cleanly classifies these because the first occurs often with other drugs-related queries, the second never does, and the last one does sometimes. For this reason, the relative distribution of queries containing \emph{high} within the positive class and the negative class mimics its meaning in reality.

\xhdr{Explainable Model} For each stages of the expansion process, we get human readable explanations. In rare cases where some seed is resulting in bad expansions, it is easy to detect this because we can work backward from an unexpected query, and each stage is readable, made up of words instead of inscrutable numbers.

\xhdr{Hybrid Serving Model for Classifiers}
When PinSets expansions are used to train a deep-learned textual model (such as fastText \cite{Joulin2016}), it performs well because the training data represents ground data well. But because textual models have trouble with non-compositional queries, the trained model can still misclassify examples from the training data. One such misclassified example from our experiment is \query{durban poison}, an obscure name for one strain of marijuana. This is rare in queries, and the fastText model does not latch on to its unsafe sense.

Despite this drawback, there is good reason to use deep-learned models in production: they are easy to deploy, and given that language is in fact compositional to a large extent, they work well, including for unseen queries. Training queries misclassified by the trained model are not numerous, and we can easily load these disagreements and thus obtain a hybrid model, combining the behavioral characteristics for frequently seen queries and fastText's generalizing capability for unseen and rarely seen queries.

\section{Location in Pinterest's Ecosystem}\label{sec:ecosystem}
\begin{figure}
	\centering
	\includegraphics[angle=90,width=0.47\textwidth]{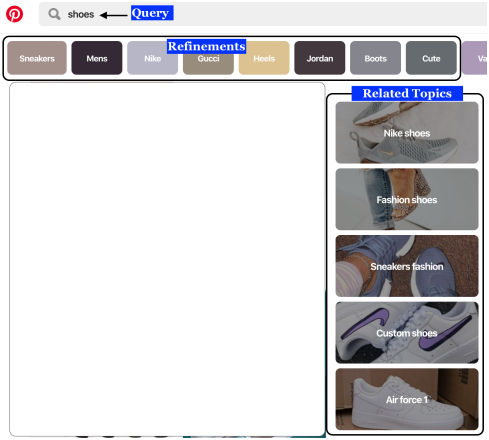}
	\caption{Examples of query suggestions at Pinterest. Apart from query completions for incomplete queries (not shown in this figure), Pinterest shows search guides containing query refinements (at the top) and related recommended stories (lower right). Queries play a role in the identification of candidate suggestions.}
	\label{fig:guides}
\end{figure}

Expanded query sets have many applications all across Pinterest. Here we consider the specific application of query suggestions, which play a role in multiple Pinterest features. These include search auto-complete (which suggests completions for a partially typed query), search guides (which suggest related topics for a given search), recommended search stories and recommended search emails. They play a key factor for improving Pinterest's popularity as they help users find search topics that may best fit into their interests. Figure \ref{fig:guides} shows some examples of query suggestions.

Each of these use cases has the potential to surface unsafe queries, and we therefore do automated detection and filtering before query recommendations from any search product are shown to the user.

\begin{figure}
	\centering
	\includegraphics[width=0.47\textwidth]{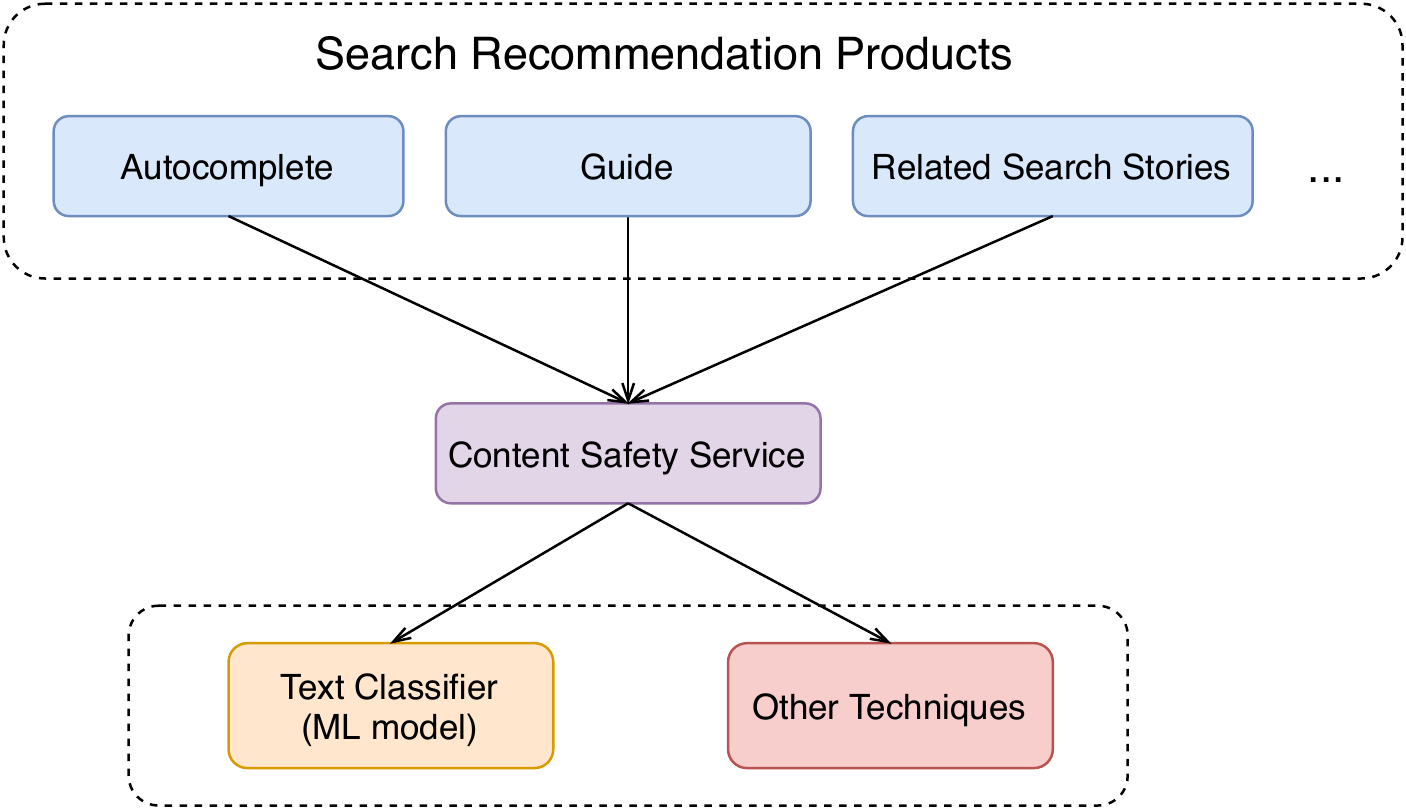}
	\caption{The Content Safety Service (purple) is the intermediary between user facing surfaces (blue) and internal details of query safety evaluation (orange and red).}
	\label{fig:flow}
\end{figure}

Pinterest has a centralized service that handles query safety for all these use cases. In Figure~\ref{fig:flow} the purple box is the central safety clearinghouse that serves the products represented by the blue boxes. Under the hood, it uses a deep learned classifier (orange box), along with other techniques necessary for blocking emerging threats that call for an urgent response (red box).

Most of the training data for the deep learned model comes from PinSets. This is augmented with some historical human rated data.

\section{How PinSets Works}

\newcommand{\seedset}{\ensuremath{\mathcal{S}}}
\newcommand{\featureset}{\ensuremath{\mathcal{F}}}
\newcommand{\expandedset}{\ensuremath{\mathcal{I}}}
\newcommand{\positiveset}{\ensuremath{\mathcal{P}}}
\newcommand{\negativeset}{\ensuremath{\mathcal{N}}}
\newcommand{\allqueries}{\ensuremath{\mathcal{Q}}}
\newcommand{\allsessions}{\ensuremath{\mathcal{C}}}
\newcommand{\bipartite}{\ensuremath{\mathcal{B}}}
\newcommand{\model}{\ensuremath{\mathcal{M}}}

Apart from the seed queries, PinSets also uses query sessions. Let \allqueries\ be the universe of queries. A single \emph{query session} $s\subset\allqueries$ is a set of queries made by a single user in a short duration, and \allsessions\ is our corpus of all sessions containing between 5 and 20 queries. For this discussion, we restrict ourselves to English query sessions, but the technique is language agnostic.  

Algorithm \ref{algo:overall} describes the expansion in broad brushstrokes. We produce the weighted bipartite graph \bipartite\ only once for the corpus \allsessions\ of sessions (details in Algorithm \ref{algo:B}). Each seed set reuses the same graph.  PinSets expands the seed queries (\seedset) in two phases. The first phase (Section \ref{sec:ph1}, Algorithm \ref{algo:p1}) expands \seedset\ to \expandedset, which only contains queries from the head and torso (i.e., queries seen in at least a hundred sessions). The second phase (Section \ref{sec:ph2}, Algorithm \ref{algo:p2}) scores all queries seen in the corpus. The positive training set \positiveset\ is those queries above a particular score threshold ($t_p$), whereas the negative training set \negativeset\ is queries below a different, lower threshold ($t_n$, with $t_n<t_p$). Queries with score in the range $[t_n, t_s]$ are not part of either sets.

\algblock{Input}{EndInput}
\algnotext{EndInput}
\algblock{HyperParameters}{EndHyperParameters}
\algnotext{EndHyperParameters}
\algblock{Output}{EndOutput}
\algnotext{EndOutput}
\newcommand{\Desc}[2]{\State \makebox[2em][l]{#1}#2}

\algrenewcommand{\algorithmiccomment}[1]{\dotfill$\triangleright$ #1}

\begin{algorithm*}[!t]
	\caption{Overall expansion algorithm.}\label{algo:overall}
	\begin{algorithmic}
		\Input
		\Desc{\allsessions}{Corpus of sessions}\Comment{Each session is a set of queries}
		\Desc{\seedset}{Seed queries to expand}
		\EndInput
		\Output
		\Desc{\positiveset}{queries strongly related to \seedset}\Comment{Used as positive examples for training classifier}
		\Desc{\negativeset}{queries strongly unrelated to \seedset}\Comment{Like above, but negative. {\em Not} the complement of \positiveset}
		\EndOutput
		\State Create bipartite graph \bipartite\Comment{Done once and reused. See Sections \ref{sec:ngrams}--\ref{sec:assoc} and Algorithm \ref{algo:B}}
		\State Expand \seedset\ to head and torso, producing \expandedset\Comment{See Section \ref{sec:ph1} and Algorithm \ref{algo:p1}}
		\State Score all queries as they relate to \expandedset \Comment{See Section \ref{sec:ph2} and Algorithm \ref{algo:p2}}
		\State Two different thresholds produce \positiveset\ and \negativeset
	\end{algorithmic}
\end{algorithm*}

\subsection{Queries, Ngrams, and Their Bipartite Graph}\label{sec:ngrams}

A \emph{query ngram} is a unigram or a bigram in a query. We na\"ively split on space to obtain ngrams, and this can be replaced with more linguistically motivated splits. A \emph{co-occurring ngram} for query $q$ in session $s$ is an ngram of some query in session $s$ such that it is not an ngram of query $q$. Thus, if a session has two queries \query{a b c} and \query{c d e}, then the co-occurring ngrams for the first query in this session are \emph{d, e, c d} and \emph{d e} but not \emph{c}.

We preprocess the corpus \allsessions\ once into a bipartite graph \bipartite\ with queries on one side and ngrams from queries on the other. In this bipartite graph, a query $q$ is connected to ngram $n$ if $n$ is strongly co-occurring with $q$. The mathematical properties of this association strength are crucial, and require their own section (Section \ref{sec:assoc}). 

Note that \bipartite\ only contains queries and ngrams seen in at least a hundred sessions, and thus represents the head and the torso of Pinterest queries.

\subsection{Preprocessing: Association Strength in \bipartite}\label{sec:assoc}

Of central importance are the weights of edges connecting a query $q$ with an ngram $n$, and our choice of association strength is driven by two concerns: 
\begin{enumerate}
	\item deal appropriately with the lopsided emphasis that any real corpus exhibits.  Real data is necessarily uneven in various ways: different languages are unevenly represented (for instance, in our data, English is far more prevalent than German which is more prevalent than Swahili) and different topics are unevenly represented (e.g., fashion and home decor are more prevalent than numismatics), and
	\item place greater faith in ngrams typically seen with a query.
\end{enumerate}

Commonly used association strength measures such as Point-wise Mutual Information (PMI) overestimate the strength between rare items \cite{Role2011}. Let $|\allsessions|$ be the number of sessions in our corpus, $|q|$ be the number of sessions with query $q$, $|n|$ be the number of sessions with ngram $n$, and $c(q, n)$ be the number of sessions where $n$ co-occurred with $q$. How frequently do we expect to see $q$ and $n$ together by chance? That number is $|n| |q| / |\allsessions|$. This estimate will be much less than 1 when both $q$ and $n$ are rare, and even a single observation of the two together will lead to a high PMI. Thus, a Swahili query and a Swahili ngram will tend to have very high association strengths arising from the rarity of these relative to the entire corpus.

Placing greater faith in ``typically seen'' can be explained by the following example. If a query $q$ is associated with two ngrams, $n_1$ and $n_2$, with $|n_1|=100, c(q, n_1)=10, |n_2|=500, c(q, n_2)=50$, then the PMI will treat both ngrams as equally strongly associated with $q$, but we would like to prefer $n_2$, which is less likely to be a statistical fluke.

We develop a new scoring function based on (1) and (2) for association strength (Equation \ref{eqnassoc}). The first component of the sum is a PMI variant independent of the corpus size $|\allsessions|$ (addressing concern 1), and the second component is a correction addressing concern 2. 

\begin{equation}
\label{eqnassoc}
w(q, n) = \log{\frac{c(q, n)^2}{|q| |n|}} + \log{\frac{c(q, n)}{|q|}}
\end{equation}

The construction of the bipartite graph \bipartite\ take one parameter: the association strength threshold $t_w$ (a suitable value is -18). If $q$ and $n$ have association strength $w(q, n)$, we retain the edge only if $w(q, n) > t_w$, and set the edge weight to $w(q, n) - t_w$, thereby making all edges have a positive weight.

\begin{algorithm*}[!t]
	\caption{Generating bipartite graph \bipartite.}\label{algo:B}
	\begin{algorithmic}
		\Input
		\Desc{\allsessions}{Corpus of sessions}
		\EndInput
		\HyperParameters
		\Desc{$t_w$}{Association strength threshold}
		\EndHyperParameters
		\Output
		\Desc{\bipartite}{bipartite graph with queries to ngrams}
		\EndOutput
		
		\State $|q|$ \Comment{Number of sessions with query $q$}
		\State $|n|$ \Comment{Number of sessions with ngram $n$}
		\State $c(q, n)$ \Comment{Number of sessions with $q$ cooccuring with $n$}
		\State $w(q, n) \gets \log{\frac{c(q, n)^2}{|q| |n|}} + \log{\frac{c(q, n)}{|q|}}$\Comment{Association strength, \ref{sec:assoc}}
		\State $a(q, n) \gets w(q, n) - t_w$ \Comment{Edge weight; $(q, n)\in\bipartite$ iff $a(q,n)>0$}
	\end{algorithmic}
\end{algorithm*}

\subsection{\bipartite: Mixing The Behavioral and The Textual}\label{sec:mixing}

The bipartite graph \bipartite\ contains, on one side, textual elements obtained by ``looking inside'' queries: these are the ngrams from queries. \bipartite\ also contains, on the other side, queries as black boxes: these are full queries whose internal structure is not inspected: it matters not if they are single words or ten word queries with exclamation marks. This mixing has benefits that should be pointed out.

\xhdr{Illustrative Example} To show the benefits, we will look at queries most closely related to the query \query{maryjane smoke}, namely, those most easily reached by a random walk of length 2 starting at that query. That is, we will look at some ngrams connected to this query, and at other queries connected to some of these ngrams. The goal here is to show how the mixed structure allows for queries with completely different words to be closely related, while at the same time be immune to ambiguity of the words in the query.

\begin{table}[t]
\centering
	\begin{tabular}{ccc}
		weed & smoke bangs & mary jane \\
		marijane&marijuana&smoke cannabis\\
		jane&weedlover&stoner girl \\
		stoner&bongs&smoker weedgirl\\
		420&blunts&medical marijuana\\
		kush&marijuanna&maryjane watson\\
	\end{tabular}
	\caption{Some ngrams associated with \query{maryjane smoke}. That is, sessions with the query \query{maryjane smoke} frequently contains these ngrams.}\label{tab:getting_f}
\end{table}

Table \ref{tab:getting_f} shows some of the top ngrams associated with the query \query{maryjane smoke}. Recall that these are ngrams seen much more than expected in sessions containing this query. This is a drugs related query, and naturally there are many drug related terms. The word \emph{maryjane} is ambiguous: it is the first name of Spider-man's friend, for instance. Most of the related ngrams are about drugs (such as the typo \emph{marijuanna}), although we also see Spider-man's friend, Maryjane Watson, with a weak connection. Although the query contains an ambiguous term, the query itself is not ambiguous, and the query's internal words have no effect on its outgoing edges; only the full query's meaning has effect, since that is what drives users to use it with other queries in a session.

Top queries associated with the ngram ``marijuanna'' are shown in Table \ref{tab:getting_q}, and are clearly drugs related. Many of the same queries are also associated with the other ngrams shown in Table \ref{tab:getting_f}, and would be reached with high likelihood by a random walk. Note the diverse vocabulary in these queries. To reemphasize, the reason these queries are associated with ngrams such as \emph{marijuanna} is because of the meaning of the full query, not of potentially ambiguous terms within. Queries associated with Spider-man are far less likely to be reached given the single shared ngram between that query and \query{maryjane smoke}.

\begin{table*}[t]
\centering
	\begin{tabular}{ccc}
		\query{weed}&\query{cannabutter recipes}&\query{edible recipes cannabutter}\\
		\query{pot}&\query{marijuanan leaf}&\query{joints and blunts}\\
		\query{mariju}&\query{marijuanan tattoo}&\query{pipes and bongs}\\
		\query{ganja}&\query{weedlover}&\query{weed brownies}\\
		\query{420}&\query{joint}&\query{420 party}\\
	\end{tabular}
	\caption{Some queries associated with ngram \emph{marijuanna}. That is, sessions with these queries often contain the ngram \emph{marijuanna}.}\label{tab:getting_q}
\end{table*}

Furthermore, note that we are capable of identifying the relatedness of two rare queries about drugs  even if each is individually found in very few sessions and the two are jointly found in no session. The other ngrams in the few sessions with these queries are the glue.

All this suggests that even a simple random walk over \bipartite\ leads to a good expansion. But we can do better, taking a page from the Category Builder algorithm \cite{Mahabal2018}, as we see next.

\newcommand{\support}{\ensuremath{\sigma}}
\subsection{Phase 1: Extend to Head and Torso}\label{sec:ph1}
In this first phase of expanding seeds \seedset\ to head-and-torso queries, we first identify a weighted set of diagnostic ngrams (\featureset) for this set of seeds. We treat the seeds as a weighted set (with all weights being 1). 

The score of an ngram $n$ depends on three factors: the weights of edges in \bipartite\ connecting \seedset\ to $n$, its precision, and its recall (these are defined below). Low precision is associated with overly generic ngrams and low recall exists when an ngram is associated with just one or two seeds, indicating that it may not represent the crux of the seed set. Given these weighted ngrams, the same technique produces a weighted set of queries (\expandedset). 

\xhdr{Hyper-parameters}\label{subsec:hyper} Three hyper-parameters control the score. Two parameters, $\rho$ and $\tau$, control for penalty for low recall and precision, respectively. The third parameter, \support, is needed as protection against lopsidedness inadvertently present in the seed set. An extreme example of such lopsidedness occurs if we are expanding a seed set of all kinds of unsafe queries at once. In such a scenario, porn queries vastly outnumber all other types, and the scores of queries in the final expansion are similarly dominated by porn. Even in a more targeted seed set, there is unintended imbalance. To account for this, when scoring ngrams, we limit the maximum number of seeds that can contribute to a score to \support, and also limit, in the calculation for an ngrams' recall, how many seeds it must be connected to for a perfect recall.

The recall for an ngram is the fraction of seeds the ngram is connected to and precision is the fraction of that ngram's neighbors that are seeds. These numbers are adjusted based on the parameter \support.

Algorithm \ref{algo:p1} lists the precise calculations for identifying ngrams. Exactly the same equations, treating \featureset\ as the seeds, produces a weighted set of queries, \expandedset.

\begin{algorithm*}
	\caption{Phase 1 Expansion. This is applied twice: to obtain weighted ngrams from seeds (Phase 1A), and obtaining queries from these ngrams (Phase 1B).}\label{algo:p1}
	\begin{algorithmic}
		\Input
		\Desc{\seedset}{Input seeds. $w(s)$ is weight of a seed}
		\Desc{\bipartite}{Bipartite graph}
		\EndInput
		\HyperParameters
		\Desc{\support}{Seed support size, default 50}
		\Desc{$\rho$}{Low recall penalty, default 3.0}
		\Desc{$\tau$}{Low precision penalty, default 0.5}
		\Desc{$t_i$}{Output score threshold (only Phase 1B), default $10^{-4}$}
		\EndHyperParameters
		\Output
		\Desc{\featureset}{Phase 1A: the top 1000 ngrams by score}
		\Desc{\expandedset}{Phase 1B: queries scoring above $t_i$}
		\EndOutput
		
		\State $N(n) \gets \{q | (q, n)\in\bipartite\}$ \Comment{neighbors of $n$}
		\State $\seedset(n) \gets N(n)\cap\seedset$ \Comment{seeds associated with $n$}
		\State $\seedset_\support(n)\subseteq\seedset(n)$ \Comment{\support\ of the  most strongly associated seeds}\\
		
		\State $r(n)\gets \dfrac{|\seedset(n)|}{|\seedset|}$ \Comment{recall for $n$}
		\State $r_\support(n) \gets \dfrac{|\seedset_\support(n)|}{\min{(|\seedset|, \support)}}$ \Comment{adjusted recall for $n$}\\
		
		\State    $p(n)\gets\dfrac{|\seedset(n)|}{|N(n)|}$\Comment{precision for $n$}
		\State    $p_\support(n)\gets\dfrac{|\seedset(n)|}{\max{(|N(n)|, \support)}}$\Comment{adjusted precision for $n$}\\
		
		\State    $u_\support(n)\gets\sum_{q\in\seedset_\support(n)}{w(q)\bipartite(q, n)}$\Comment{uncorrected score for $n$}
		\State    $a_\support(n)\gets u_\support(n)r_\support(n)^\rho p_\support(n)^\tau$\Comment{final score for $n$}
		
	\end{algorithmic}
\end{algorithm*}

\subsection{Phase 2: Extend Deep into the Tail}\label{sec:ph2}
The second phase is purely behavioral. For each session $s\in\allsessions$, we count how many queries from \expandedset\ are present, $|s\cap\expandedset|$. For each unique query $q$, we count the following:
\begin{itemize}
	\item $t$ is the number of sessions containing query $q$.
	\item $u$ is the number of unsafe sessions containing $q$. For a session $s$ to be unsafe we need to see \emph{three other} unsafe queries (i.e., if $q\in\expandedset$, we require $|s\cap\expandedset| \ge 4$ otherwise we require $|s\cap\expandedset| \ge 3$.
	\item The score for the query is a smoothed version of the ratio $u/t$. Specifically, we use $(u+1)/(t+30)$ to prevent spurious high scores for rare queries.
\end{itemize}

The positive training data \positiveset\ consists of queries seen in at least 10 sessions with a score of at least 0.1, and the negative training data \negativeset\ is made up of queries seen in at least 300 sessions with a score below 0.005.

\begin{algorithm*}
	\caption{Phase 2 Expansion.}\label{algo:p2}
	\begin{algorithmic}
		\Input
		\Desc{\allsessions}{Corpus of sessions}
		\Desc{\expandedset}{Queries identified in Phase 1}
		\EndInput
		\HyperParameters
		\Desc{$t_p$}{Score threshold for positive data}
		\Desc{$t_n$}{Score threshold for negative data}
		\EndHyperParameters
		\Output
		\Desc{\positiveset}{queries strongly related to \seedset}
		\Desc{\negativeset}{queries strongly unrelated to \seedset}
		\EndOutput
		
		\State $|q|\gets |\{s|q\in s\}|$\Comment{Number of sessions with $q$}
		\If{$q\in\expandedset$}
		\State $u(q)\gets |\{s|q\in s, |s\cap\expandedset|\ge 4\}|$\Comment{Sessions with 3 other unsafe}
		\Else
		\State $u(q)\gets |\{s|q\in s, |s\cap\expandedset|\ge 3\}|$\Comment{Sessions with 3 other unsafe}
		\EndIf
		\State $a(q)\gets \dfrac{u(q)+1}{|q|+30}$\Comment{smoothed unsafe session fraction}
		\State \positiveset$\gets\{q|a(q) \ge t_p\}$\Comment{Above threshold is positive}
		\State \negativeset$\gets\{q|a(q) \le t_n\}$\Comment{Below another threshold is negative}
		
	\end{algorithmic}
\end{algorithm*}

\subsection{Deep Learning and Hybrid Serving}
The sets \positiveset\ and \negativeset\ are used to train a deep-learned textual model \model\ that can evaluate the safety of seen and unseen queries. It is a textual solution, and naturally limited in its ability to handle non-compositional queries. We therefore identify training examples where the model disagrees with the behavioral data, and this relatively small number of queries is easy to load and use in deciding query safety.

\section{Experiment: Expanding Drug Seeds}

\newcommand{\intermediate}{\textsc{Intermediate}}
\newcommand{\drugs}{\textsc{Drugs}}
\newcommand{\nondrugs}{\textsc{Non-Drugs}}
\newcommand{\intermediaterated}{\textsc{Intermediate-Rated}}
\newcommand{\drugsrated}{\textsc{Drugs-Rated}}
\newcommand{\nondrugsrated}{\textsc{Non-Drugs-Rated}}

We begin with 20 seed queries concerning drugs (shown in Table \ref{tab:seeds}). The two phase expansion process produced 4,711 head-and-torso queries (\intermediate) and expanding these lead to 15,670 drugs-related queries (\drugs) and 1.7 million queries not about drugs (\nondrugs). We used human raters to evaluate the quality of \intermediate, \drugs, and \nondrugs. For each of those three sets, we sampled 2,000 queries (\intermediaterated, \drugsrated, and \nondrugsrated) for human rating.

Further, we trained a fastText model, with the positive examples being ($\drugs-\drugsrated$) and the negative examples sampled from ($\nondrugs-\nondrugsrated$), and looked at the errors made by this model. 

\begin{table*}
	\centering
	\begin{tabular}{cc}
		\query{marijuana pills} & \query{cannabis black and white} \\ 
		\query{drugs meth crystals} & \query{weed be good together}\\
		\query{weed jewlery} & \query{magic mushshroom psychedelic} \\
		\query{ketamine sniff} & \query{cbd lemonade} \\
		\query{canabis indica} & \query{growing marajuana beginners seeds} \\
		\query{meth head} & \query{baked stoner} \\
		\query{cannabis sativa oil} & \query{mdma aesthetic} \\
		\query{box of weed} & \query{canabis growing outdoor} \\
		\query{weed for pain} & \query{marijuanna colors} \\
		\query{maryjane smoke} & \query{marijuanna edible recipes cannabis oil}\\
	\end{tabular}
	\caption{(Experiment) 20 drugs-related seed queries.}
	\label{tab:seeds}
\end{table*}

\subsection{Evaluation Methodology}

The three evaluation sets of 2,000 queries each (i.e, \intermediaterated, \drugsrated, and \nondrugsrated) were sent to human raters. Raters used the Pinterest standard format for evaluating safety, classifying each query as either safe or assigning it to one of several unsafe categories, one of which is Drugs. 

Each query was rated by five raters, and we used majority rating to assign categories to queries. if a query gets ``Drugs'' as the majority vote, we consider it to be about drugs. Note that non-drugs can still be unsafe for other reasons (such as porn or graphic violence). 

\subsection{Phase 1 Expansion}

Phase 1 expansion identified 210 diagnostic ngrams (Table \ref{tab:featureset}) and an expanded set of 4,711 head-and-torso queries (Table \ref{tab:expanded}).

\xhdr{Diagnostic ngrams (\featureset)} One of the top scoring diagnostic ngrams is \emph{marijuana}, which is associated with six of the seed queries, and gets a final score of 0.01. This low score is caused by the small seed set and the fact that \emph{marijuana} is connected to over 6,000 queries, giving it a low perceived precision. More precise ngrams (i.e., with fewer non-seed queries) include \emph{methanphetamines}, which is connected to just one seed but also just 56 queries in all. We also see more generic ngrams. One example is \emph{quotes}, which is connected to a single seed and about half a million queries overall, resulting in the very tiny score of $1.3*10^{-8}$. If we had started with a hundred seeds instead of with just 20, these generic features get dropped when we keep the top thousand ngrams.

\begin{table}
\centering
	\begin{tabular}{ccc}
		cannabis sativa & stoner & sativa \\
		pain medical & cannabis & drugs \\
		sativa plants & marijuana & weed \\
		beginners outdoor&drug&lsd\\
		maryjane tattoo&mdma molly&of meth
	\end{tabular}
	\caption{(Experiment) Some of the 210 diagnostic ngrams associated with the 20 seed queries in Table \ref{tab:seeds}.}
	\label{tab:featureset}
\end{table}

\xhdr{Phase 1 Expansion (\intermediate)}
4,711 head and torso\linebreak queries were observed with a score above $10^{-4}$, and some of these are shown in Table \ref{tab:expanded}. One of the highest scoring queries is \query{smoke weed everyday}, which is associated with a total of 177 ngrams, 35 of which are in \featureset. Even a relatively obscure, typo-ridden query such as \query{easy marijuanna edible recipes} is associated with 19 ngrams in \featureset and thus scores highly in the expansion. The queries \query{spider man maryjane} and \query{j scott campbell spiderman} end up with a score below $10^{-8}$, much below the threshold, and are thus not part of the 4,711 queries.

\begin{table}
	\centering
	\begin{tabular}{cc}
		\query{addiction photography}&\query{art weed}\\
		\query{420 aesthetic}&\query{estacy}\\
		\query{edible recipes cannibus} & \query{lad} \\
		\query{marijuana pipes} & \query{stoner decor} \\
		\query{weed pipes} & {70s weed}\\
		\query{\#indica}&\query{marihuana art}\\
		\query{stoner tattoo}&\query{baddies smoking}\\
		\query{hemp oil}&\query{drug addict aesthetic}\\
		\query{smoking pipe bowls}&\query{bong tattoo}\\
	\end{tabular}
	\caption{(Experiment) A few of the 4,711 queries obtained in Phase I by expanding the 20 seeds.}
	\label{tab:expanded}
\end{table}

\xhdr{Human evaluation of \intermediaterated} 
These 2,000\linebreak queries are expected to be drugs related, and 97.9\% of them indeed get a majority label of ``Drugs'', as seen in Table \ref{tab:accuracy}. It is instructive to look at the 42 queries that don't get a majority label of ``Drugs'' (Table \ref{tab:expandederrors}).

\begin{table}
\centering
	\begin{tabular}{ccc}
		\query{addict}&\query{grower}&\query{chillum}\\
		\query{drigs}&\query{drogs}&\query{ganza}\\
		\query{maconh}&\query{maconhaa}&\query{maconharia}\\
		\query{edinles}&\query{ruderalis}&\query{hasis}\\
		\query{red eyes}&\query{smoking}&\query{alcohol}\\
	\end{tabular}
	\caption{(Experiment) Queries in Phase I expansion without ``Drugs'' as the majority human rating. Some of the 42 such queries (out of the 2,000 rated) are shown. Note the several typos for drugs and foreign terms for drugs. The last row drifts the most from drugs.}
	\label{tab:expandederrors}
\end{table}

Several of these queries are short and typos (such as \query{ganza}, \query{hasis}, and \query{drigs}, which are likely typos for \emph{ganja, hashish, } and \emph{drugs}), and more importantly, these are used by users in the same session as drug related terms, and thus may lead to drugs-related content despite typos.

Others are foreign language of technical terms, such as \query{chillum} and \query{ruderalis}, the latter being the botanical name of one strain of cannabis.

The final row in the table displays more problematic errors: things that perhaps co-occur with drugs but represent a semantic drift and are certainly not, in and of themselves, drugs-related. We see in the next subsection how these errors get cleaned up by phase 2. 

\subsection{Phase 2 Expansion}
The expansion resulted in 15,670 unique queries in \drugs\ and around 1.7 million unique queries in \nondrugs.
The \nondrugs\ examples span the gamut of Pinterest queries.

\xhdr{Human Evaluation} 2,000 queries were sampled from \nondrugs. Not a single query was classified as drugs-related, implying 100\% purity of this sample. 2,000 queries were sampled from \drugs, and 1986 of this had a majority rating drugs, representing a 99.3\% purity of this sample. Table \ref{tab:mislabeled} lists the 14 queries where majority vote was not ``drugs'', and even these are close to being drugs-related. Some are foreign language queries about drugs, other typos for drugs-related queries, and some about the tobacco weed used to roll marijuana blunts. The lower 6 queries represent the greatest semantic drift, and concern vaping, psychedelic art, and the medicine Zantac.

\begin{table}
	\centering
\begin{tabular}{crcr}
		&\textbf{\# Queries}&\textbf{PinSets}&\textbf{fastText}\\
		\intermediate&4,771&97.9\%&92.5\%\\
		\drugs&15,670&99.3\%&95\%\\
		\nondrugs&1,735,286&100.0\%&100\%\\
	\end{tabular}
	\caption{(Experiment) Human evaluation of expansion based on majority vote among five raters. The rows correspond to Phase 1 expansion (\intermediate), and the positive and negative output of Phase 2 (\drugs\ and \nondrugs). In each case, 2,000 queries were sent to raters. The last two columns report precision for PinSets and a fastText model (Section \ref{sec:expt_fasttext}).}
	\label{tab:accuracy}
\end{table}

\begin{table*}
\centering
	\begin{tabular}{cc}
		\textbf{Queries}&\textbf{Comments}\\\hline
		\query{droga}, \query{drogue}&Spanish and French for drug\\
		\query{maconhaa papeis de parede}&Spanish, ``Marijuana wallpaper''\\
		\query{metg}, \query{tripy}, \query{wed}&Typo for \em{meth, trippy}, and \em{weed}\\
		\query{backwood arts}&tobacco leaf for rolling weed\\
		\query{backwood tumblr}&tobacco leaf for rolling weed\\\hline
		\query{cute wax pen}, \query{pen battery}&concerns vaping\\
		\query{429}&inscription on zantac\\
		3 queries re: psychedelic art&3 such queries
	\end{tabular}
	\caption{(Experiment) These 14 queries (out of 2000 rated) did not get ``Drugs'' as majority vote. The upper 8 are arguably drugs-related, and even the lower 6 are in the semantic vicinity.}
	\label{tab:mislabeled}
\end{table*}

\subsection{Classifier Trained on Phase 2 Output}\label{sec:expt_fasttext}
As previously remarked, a textual classifier has trouble with compositional queries. We trained a fastText classifier with phase 2 output (but not using those queries sent to human raters). Human evaluation of the classifier revealed a 95\% precision on \drugs, although it had a perfect score on \nondrugs. The misclassified 5\% has several non-compositional queries that reuse common words in unusual ways, for example, \emph{Durban poison} (a South African marijuana variety) and \emph{Moon Rock bud} (cannabis buds dipped in hash). Table \ref{tab:delta} shows some of these queries.

\begin{table*}[t]
	\centering
	\begin{tabular}{cc}
		\textbf{Queries}&\textbf{Comments}\\\hline
		weex; estacy; mary hane; kusg; acud; mary jame; schrooms&typos\\
		molly; methed~up; marwana&slang\\
		durban poison; moon rock bud; supercropping&exotic slang?\\
		snort coke&The term {\em Coke}, for Coca Cola, is usually safe
	\end{tabular}
	\caption{(Experiment) Some unsafe queries misclassified as non-drugs by fastText.}
	\label{tab:delta}
\end{table*}

\subsection{Handling Ambiguity}

Both \emph{weed} and \emph{pot} are ambiguous terms, and typically safe in Pinterest query streams. How well do we tell apart the needles (i.e., unsafe uses of these terms) from the haystack? A query with \emph{pot} is 250 times likelier to be in the safe training data than in unsafe. PinSets' behavioral component makes it robust to the textual ambiguity represented by the word \emph{pot}, resulting in highly accurate training data, as can be observed from Table \ref{tab:pot}. \emph{Weed} has the same story, as seen in Table \ref{tab:weed}.

\begin{table*}[t]
	\centering
	\begin{tabular}{cc}
		\textbf{In \nondrugs}&\textbf{In \drugs}\\\hline
		\query{soups in a crock pot} & \query{edibles pot candy}\\
		\query{flower pot ideas}&\query{pot cookies weed recipes}\\
		\query{one pot pasta}&\query{grow room pot}\\
		\query{pot luck ideas}&\query{smoking tips pot}\\
		\query{liquid smoke instant pot}&\query{pot smoke}\\
	\end{tabular}
	\caption{(Experiment) Safe and unsafe queries containing \emph{pot}, as classified by PinSets. Among unsafe uses, we saw 278 unique queries asked 49K times, where as safe uses were 6K unique queries asked 13 million times. The last row shows that even the joint presence of \emph{smoke} and \label{tab:pot}
	\emph{pot} doesn't necessarily render a query unsafe.}
\end{table*}

\begin{table*}[t]
	\centering
	\begin{tabular}{cc}
		\textbf{In \nondrugs}&\textbf{In \drugs}\\\hline
		\query{butterfly weed}& \query{weedgirls stoner girl smoke weed}\\
		\query{diy weed killer}& \query{buy weed medical marijuana}\\
		\query{horny goat weed}& \query{badass weed wallpaper}\\
		\query{barrett wilbert weed}& \query{smoke weed tricks} \\
		\query{sea weed}&\query{buy weed how to}\\
		\query{weed identification}&\query{ganjaa wallpapers smoking weed}\\
		\query{tumble weed}&\query{420 humor smoking weed}\\
	\end{tabular}
	\caption{(Experiment) Safe and unsafe queries containing \emph{weed}, as classified by PinSets. }\label{tab:weed}
\end{table*}

\section{Effect on Pinterest Query Recommendation Safety}
Making search recommendation safe is one of the top priorities of Pinterest. In an earlier iteration (before PinSets), we used a fastText model trained on human labeled queries and on unsafe query expansions reported by users. 

We regularly monitor the safety of suggested queries by sampling queries and getting these rated by human raters. We monitor unsafe suggestions in all our unsafe categories (Drugs, Porn, etc), and can thus evaluate the impact of model changes on overall safety.

We used PinSets to generate additional training data. Starting with known unsafe seed queries in each unsafe domain, we produced a large, scored collection of unsafe queries. Top 5000 of these were sent to human raters, and these were 97\% accurate. The human raters also assigned categories to the expanded queries, and these clean queries were used as new seeds to identify over a million unique unsafe queries. These newly identified queries form the bulk of overall training data, accounting for over 80\% of unique queries.

When we augmented the training data with these additional queries generated by PinSets, our routine measurements showed a 90\% drop in unsafe suggestions. Some domains with very large vocabularies but relative rarity in the query stream, such as drugs, saw even larger drops.

One danger with such a high drop in unsafe query suggestion is the potential for collateral damage: maybe we are removing safe suggestions as well as unsafe, and this will show up as, for instance, a drop in certain metrics of user engagement. We also monitor these metrics, and the launch of the improved model was neutral (with under 1\%), suggesting that safe queries were not harmed.

\subsection{Hybrid Serving}
Although the deep learned system has been trained on data obtained using queries' historical engagement, it is itself a purely textual model basing its decision only on the query text. For non-compositional queries, the fastText model is less accurate than the training data used.

If we treat the training data as golden, we observe the following. Among known unsafe queries, fastText is quite accurate: it classifies 99.12\% queries correctly. Among known safe, it clocks in at 97\% correct, and thus errs on the side of caution.

Misclassified queries include those with typos (intentional or otherwise): \query{\#c\"um}, \query{death quoats}, \query{panic attach relief}, \query{b\"o\"oty,} and \query{c\o ck}. Among false positives, we see examples with a typically unsafe term: \query{nailart nude shape}, \query{hot wheels cake 4}, \query{crockery unit cum bar}, \query{beat up truck}.

A class of queries that is hard to handle compositionally concerns an area Pinterest deeply cares about: mental health. Queries that co-occur with other queries clearly about sadness or depression are often hard to fathom as a combination of their individual terms: \query{there is no hope} and \query{its over}.

For queries seen frequently, we can be confident of the safety or lack thereof, and thus trust such data more, allowing it to trump the fastText model for these queries. We need not load all these queries, however: if need only to store cases where the two models disagree, and this is small enough to be loaded without impacting latency. 

\section{Related Work}

\xhdr{Set Expansion} Set Expansion is the well studied problem of expanding a given set of terms by finding other semantically related terms \cite{Wang2007,Shen2017}. Our Phase 1 expansion is based most closely on our prior work, namely the Category Builder algorithm \cite{Mahabal2018}, but with additional innovations like accounting for imbalanced seed emphasis via the $\tau$ hyper-parameter. The mixing of textual and behavioral (which gives protection against ambiguity but still gets generalization benefits of compositionality) is novel.

\xhdr{Expanding query-sets to generate training data} Mining an unlabeled corpus of queries to augment a small known set of labeled examples with a much larger set of pseudo-labeled examples is a form of semi-supervised learning. This is often treated as label propagation via graph transduction on a homogeneous graph \cite{Zhu2005, Elezi2018}. Phase 1 of our approach uses a heterogeneous graph (with queries and ngrams), with the ensuing benefits already noted.

Another formulation often used is that of classification for expansion, where the limited labeled data is used to train a classifier which is then used to classify the unlabeled data, resulting in much larger training data for the final model. This approach is used by \cite{Wulczyn2017} to classify personal attacks, and they expand 100K human labeled comments to 63M machine labeled. The classifier is textual, however, and ambiguous terms that the original classifier did not learn well will be mishandled in the expansion, something PinSets does not suffer from, and of course such an approach is a non-starter if we start with only a few dozen seeds.

\xhdr{Unsafe Text Identification} Much research in this domain concerns machine learning models appropriate for identifying unsafe text \cite{Yenala2018,Zhang2018,Xiang2012}. As such, this line of work is orthogonal to the present work and the two can be used jointly: we produce training data that can be used by these models.

A large fraction of unsafe text identification work concerns longer forms of text, such as tweets and messages (e.g, \cite{Pant2019,Yenala2018,Wulczyn2017,Xiang2012,Zhang2018}). These longer forms have been noted to be easier since they offer more context \cite{Yenala2018}, which a short query such as \query{c\o ck} clearly doesn't. Our black-box approach infers context from the session rather than from within the query
, allowing us to maintain high precision when classifying shorter text.

\section{Conclusions and Future Work}
The immense diversity of vocabulary in a domain can be accessed by a combination of techniques, utilizing the topic stickiness of sessions and exploiting the fact that a query is not ambiguous to the user issuing it and that we can exploit the user's behaviour, as revealed by their word choice in other queries they issue just before or after.

The ideas presented here are language agnostic with one English-specific exception that won't generalize well to languages such as Thai: the na\"ive, white-space-based tokenization. Future work should ensure applicability to a wider range of languages.

Although we focused here on query safety, the mechanisms are widely applicable, and in subsequent work we are looking into enriching queries associated with taxonomies of user interests.

\section*{Acknowledgements}
We would like to thank Chuck Rosenberg for his many helpful suggestions.

\bibliography{naaclhlt2019}
\bibliographystyle{acl_natbib}

\end{document}